\documentclass[sigconf]{acmart}
\usepackage{multirow}
\AtBeginDocument{%
  \providecommand\BibTeX{{%
    \normalfont B\kern-0.5em{\scshape i\kern-0.25em b}\kern-0.8em\TeX}}}

\setcopyright{acmcopyright}
\copyrightyear{2020}
\acmYear{2020}
\acmDOI{10.1145/xxxxxxx.1122456}

\acmConference[Xxxx '20]{Xxxx '20: ACM Symposium on XXX}{June xx--xx, 2020}{Xxxx, XX}
\acmBooktitle{xxx '20: ACM Symposium on XXX,
  June xx--xx, 2020, Xxxx, XX}
\acmPrice{15.00}
\acmISBN{978-1-4503-XXXX-X/18/06}

\begin{document}

\title{Refined Gate: A Simple and Effective Gating Mechanism for Recurrent Units}

\author{Zhanzhan Cheng$^{12*}$, Yunlu Xu$^{2*}$, Mingjian Chen$^2$, Yu Qiao$^3$, Shiliang Pu$^2$, Yi Niu$^{2}$, Fei Wu$^1$}
\thanks{\textsuperscript{*}Both authors contributed equally to this research.}
\affiliation{%
  \textsuperscript{1}\institution{Zhejiang University, Hangzhou, China;~~\textsuperscript{2}Hikvision Research Institute, China}
}
\affiliation{
  \textsuperscript{3}\institution{Shenzhen Institutes of Advanced Technology, Chinese Academy of Sciences, China}
}

\begin{abstract}
  Recurrent neural network (RNN) has been widely studied in sequence learning tasks, while the mainstream models (\emph{e.g.}, LSTM and GRU) rely on the gating mechanism (in control of how information flows between hidden states).
  {
  However, the \emph{vanilla} gates in RNN (\emph{e.g.}, the input gate in LSTM) suffer from the problem of \emph{gate undertraining}, which can be caused by various factors, such as the saturating activation functions, the gate layouts (\emph{e.g.}, the gate number and gating functions), or even the suboptimal memory state \textit{etc.}.
  Those may result in failures of learning gating switch roles and thus the weak performance.
  }
  In this paper, we propose a new gating mechanism within general gated recurrent neural networks to handle this issue.
  Specifically,
  the proposed gates directly short connect the extracted input features to the outputs of \emph{vanilla} gates, denoted as refined gates. 
  The refining mechanism allows enhancing gradient back-propagation as well as extending the gating activation scope,
  which can guide RNN to reach possibly deeper minima.
  We verify the proposed gating mechanism on three popular types of gated RNNs including LSTM, GRU and MGU. Extensive experiments on 3 synthetic tasks, 3 language modeling tasks and 5 scene text recognition benchmarks demonstrate the effectiveness of our method.
\end{abstract}

\maketitle

\section{Introduction}
Recurrent neural networks (RNN) receive extensive research interests because of their powerful ability to handle sequential data in various applications such as action recognition\cite{donahue2015long-term}, image captioning \cite{xu2016show}, text recognition \cite{cheng2017focus}, language translation
\cite{bahdanau2016end},
and  speech recognition \cite{graves2013speech}, {etc}. 
Specifically, the gated recurrent neural networks (GRNN) including Long Short Term Memory (LSTM) \cite{hochreiter1997long}, Gated Recurrent Unit (GRU) \cite{cho2014learning} and Minimal Gated Unit (MGU) \cite{zhou2016minimal} are the prevailing variants of RNN, which can successfully learn the long-term dependency attributing to their \emph{gates} on input $x$ and recurrent hidden $h$ (i.e., the \emph{forget/input/output} gates in LSTM, the \emph{update/reset} gates in GRU and the only \emph{forget} gate in MGU).

Then, some recent works \cite{chung2014empirical,greff2017lstm,jozefowicz2015empirical,schrimpf2017flexible} were proposed to investigate the \emph{vanilla} gating mechanism. For example,
\cite{greff2017lstm} and \cite{jozefowicz2015empirical} verified the effectiveness of these gates by analyzing different variants of GRNN such as 8 variants \cite{greff2017lstm} for LSTM and 3 variants \cite{jozefowicz2015empirical} for GRU, and expected to find out better variants of GRNN.
Unfortunately, all variants didn't obviously outperform the standard GRNN models.
\cite{schrimpf2017flexible} designed an Architecture Searcher to search potential RNN architectures, but the best searched architecture didn't follow human intuition and cannot provide clear suggestions for the design of RNN structures.

\begin{figure*}[t]
\begin{center}
   \includegraphics[width=1.\linewidth]{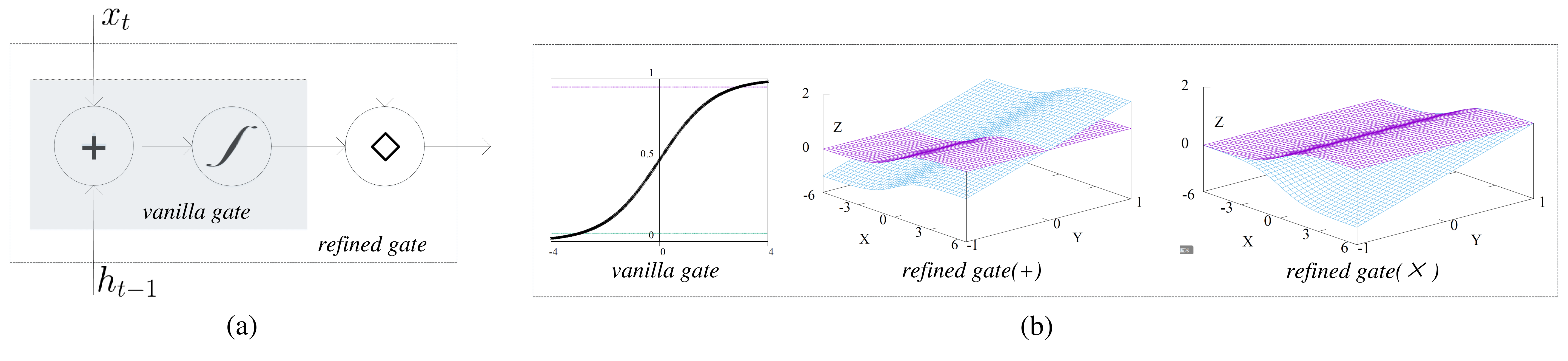}
\end{center}
   \caption{The illustration of refined gate. (a) is the refined gate block where $\int$ and $\diamond$ separately mean the activation function and the refining operation. In (b), the first sub-figure is the \emph{vanilla} gate, the second and the last are the new gates with two refining operations `+' and `$\times$'.
   In refined gates, the X, Y and Z axes correspond to the transformed $(x, h)$, the $x_t$ and the gate output, respectively.
   The purple and blue surfaces separately denote the response of \emph{vanilla} gate and refined gates.
   }
\label{fig:gate}
\end{figure*}

Empirically, we find that \emph{vanilla} gates in GRNNs tend to suffer from the problem of {insensitive gating activation {
(\textit{e.g.}, narrowing activation scope and approximately activating values with small standard deviation).}}
It means the gates can't play the switch roles very well and have limited ability to control the information flows between hidden states. We call this issue \emph{gate undertraining}.
{
\emph{Gate undertraining} problem can be derived from three perspectives:
(1) The \emph{vanilla} gating functions in popular GRNNs are always represented by the sigmoid function $\sigma$ (See the \emph{vanilla} gate part of Figure \ref{fig:gate} (b)), while the saturation characteristics of $\sigma$ limits the gate roles as addressed in early works \cite{goh2003recurrent,trentin2001networks}.
It leads to the imbalance in training of the gates (slower learning) and {the following feedforward neural networks} (faster learning).
When gates are trapped into the inevitable saturation, it means with parameter perturbed, the change to the gates tend to be small due to the sigmoid operator. Gates become persistently indiscriminate on flows while the remained network part keep tuning in a large speed.
Thereby, gates in GRNN themselves might miss the befitting learning stage.
(2) Different gate layouts (\emph{e.g.}, the gate number and the gating functions) affects the relation representation between hidden states.
For example, MGU with a minimal $\sigma$ gate may have limited ability to learn the long-dependency relationship between hidden states, because the minimal \emph{vanilla} gate undertake multiple gate roles such as the output and the forget switch roles, which requires the minimal gate more discriminative.
(3) The hidden state is responsible for dynamically holding the past and current information.
The memory cell should be able to perceive the subtle differences between the current inputs and the past states, which means the gates should be very sensitive to capture flow change.

Above issues motivate us to develop a more discriminative gating mechanism to enhance the gate control ability, which not only avoid the saturation problem, but also perceive the relationship among information flows sensitively.
And then the stronger gates can remedy the weak performance caused by undertraining. 
}
{
Inspired by previous researches of residual network \cite{he2016deep} and highway network \cite{srivastava2015training}, we develop a simple yet effective gating mechanism named refined gate, introducing a path from the input features to the outputs of nonlinear activation, as shown in Figure \ref{fig:gate}(a).
By this way, the refined gate become boundless and then eliminate the saturation problem (See the refined parts of Figure \ref{fig:gate}).
On the other hand, such shortcut paths achieve the identity mapping between the memory state and the input, which plays the differential amplifier role to better depict the switch effects.
{
Note that, different from introducing residual or highway block into RNNs for overcoming the gradient vanishing through multiple RNN layers \cite{kim2017residual,zhang2016highway}, our focus is that the refined gate 
can learn to play the pure gating roles sensitively inside the recurrent units.}
}

Concretely, instead of all the \emph{vanilla} gates implemented with a sigmoid function, i.e., $g_{vanilla}= \sigma(x, h)$, we design the refined gate denoted as $g_{refine}{=} \sigma(x, h) \diamond x$.
Here, $\diamond$ is the element-wise refining operation, which combines the activation function $\sigma$ and $x$ to enhance the traditional gating mechanism.
In this paper, we simply illustrate $\diamond$ as two modes: $+$ or $\times$ , and treat the whole refined structure as a new gate for better controlling information flows.
{
We note that 
the refined gates can provide broader and more dispersive activation scope than \emph{vanilla} gates, as shown in Figure \ref{fig:gate}(b), which is also boundless but without the saturation problem. We demonstrate that the refined gating structure can be well equipped to general GRNN units boosting performance without introducing any extra parameters as well as any vanishing and exploding gradient problems.
}

The contributions of this paper are:
(1) We provide a deeper understanding of gating mechanism in GRNNs and focus on the widely existing chanllenging problem: \emph{gate} \emph{undertraining}.
(2) We propose a new gating mechanism enhancing the vanilla gates using simple yet effective refining operations, which is verified to be well adapted in existing GRNN units like LSTM, GRU and MGU. 
(3) {We show intuitive evaluation on gate controlling ability through well-designed sequential tasks, \textit{i.e.}, adding and counting, and offer reasonable illustrations both qualitatively and quantitatively.}
(4) Experiments on various tasks, including 3 synthetic datasets 
and multiple real-world datasets (3 language modeling tasks and 5 scene text recognition benchmarks) demonstrate that the proposed gate refinement mechanism can effectively boost GRNN learning.

\section{Background}

\subsection{Gated RNNs}
{The simple recurrent architecture is hard to train properly in practice because of the vanishing and exploding gradient problems\cite{pascanu2013difficulty}.
Therefore, various gated RNNs are developed  for capturing long-term temporal dependencies, which are usually implemented by introducing various gates to control how information flows in RNN.}
We here briefly list three prevailing gated RNNs including LSTM, GRU and MGU in Eq. \ref{LSTM}-\ref{MGU}.
\begin{subequations} \label{LSTM}
\begin{align}
f_{t} &{=} \sigma(W_fx_{t}{+}U_fh_{t{-}1}{+}b_f),  \\
i_{t} &{=} \sigma(W_ix_{t}{+}U_ih_{t{-}1}{+}b_i)\label{1b}, \\
o_{t} &{=} \sigma(W_ox_{t}{+}U_oh_{t{-}1}{+}b_o)\label{1c},\\
\tilde{c}_{t} &{=} \phi(W_cx_{t}{+}U_ch_{t{-}1}{+}b_c),\label{ct} \\
c_{t} &{=} f_t\odot c_{t{-}1}{+}i_t\odot \tilde{c}_{t},\label{1e} \\
h_{t} &{=} o_t\odot\tanh(c_t).
\end{align}
\end{subequations}
\begin{subequations}\label{GRU}
\begin{align}
z_{t} &{=} \sigma(W_zx_{t}{+}U_zh_{t{-}1}{+}b_z), \\
r_{t} &{=} \sigma(W_rx_{t}{+}U_rh_{t{-}1}{+}b_r),\label{2b} \\
A&{=}r_{t}\odot h_{t{-}1},\\
\tilde{h}_{t} &{=} \phi(W_hA{+}U_hx_{t{-}1}{+}b_h), \\
h_{t} &{=}z_t\odot h_{t{-}1} {+} (1{-}z_{t})\odot \tilde{h}_t.\label{2e}
\end{align}
\end{subequations}
\begin{subequations}\label{MGU}
\begin{align}
f_{t} &{=} \sigma(W_fx_{t}{+}U_fh_{t\text{-}1}{+}b_f), \\
A&{=}f_{t}\odot h_{t{-}1},\label{3b}\\
\tilde{h}_{t} &{=} \phi(W_hA{+}U_hx_{t{-}1}{+}b_h), \\
h_{t} &{=}(1{-}f_t)\odot h_{t{-}1} {+} f_t\odot \tilde{h}_t.\label{3d}
\end{align}
\end{subequations}

The typical \emph{LSTM} \cite{hochreiter1997long} solved the gradient issues by introducing three gates (i.e., the \emph{forget} gate, the \emph{input} gate and the \emph{output} gate) to control how information flows in RNN, formalized as Eq. \ref{LSTM} where $\phi$ denotes the $\tanh$ activation.
Different from LSTM, \emph{GRU} \cite{cho2014learning} discards the output gate as well as memory state, and makes each recurrent unit to adaptively capture dependencies of different time scales by introducing the update gate and the reset gate, which is formalized as Eq. \ref{GRU}.
\emph{MGU} \cite{zhou2016minimal} is a recent proposed gated RNN with only one forget gate (shown as Eq. \ref{MGU}), which is implemented by coupling the update and reset gate of GRU into a forget gate.

In addition, Greff \emph{et al.} \cite{greff2017lstm} explored 8 variants of the LSTM architecture to evaluate the effects of these gates, and Jozefowicz \cite{jozefowicz2015empirical} evaluated 3 variants of GRU, but the results showed that none of variants can obviously outperform the standard models. Chung \emph{et al.} \cite{chung2014empirical} also compared the performance of LSTM and GRU on multiple tasks and showed the similar performance.

\subsection{Gate Functions}
{
Some nonlinear activation functions like sigmoid and tanh have been extensively used in feedforward neural networks with the inevitable \emph{gate undertraining} problem.}
Early works \cite{goh2003recurrent,trentin2001networks} attempted to relieve the saturation problem by introducing the adaptive amplitude of activation strategy. 
When it comes to GRNNs, bounded activating gates still dominate in the literature to avoid exploding gradients while such  gates fall into the \emph{gate undertraining} problem, which might result in the poor performance on various sequence data. In common, dominating gated RNNs have the same gating modality:
$sigmoid(W_g h_{t-1} +  U_g x_t + b_g)$,
where $[W_g, U_g, b_g]$ are the learnable parameters to determine the gating states.

{Some methods have been designed for allievating one of the \emph{gate undertraining} problem: the satuating state.}
\cite{gulcehre2016noisy} injected noise to \emph{sigmoid} and \emph{tanh} and replaced the soft-saturating with the proposed hard-activations.
\cite{le2015simple} proposed the rectified linear unit (ReLU) in companion with identity weight matrix initialization, which seemingly sidestepped the saturating sigmoid yet brought new problem of the Dead ReLU.
Recently, \cite{chandar2019towards} proposed a new recurrent unit without saturating gates and evaluated with vanilla RNNs on multiple tasks.
Existing works  except \cite{gulcehre2016noisy} relieved the saturating problem by designing an entirely new recurrent unit forgoing the bounded non-linear gates.
{While we devote to further excavate the ON/OFF control ability of gates in original structures themselves with only a minor yet universally applicable revision.}

\subsection{Shortcut Structures}
The residual structures \cite{he2016deep} 
and highway networks \cite{srivastava2015training} 
are influential implementations of the shortcut path for deep neural networks,
both of which provide direct path of data flows in order to simplify the model training.
As a result, works \cite{kim2017residual,lei2018simple,zhang2016highway} attempted to focus on the adaption of shortcuts into the recurrent networks.
Highway LSTM \cite{zhang2016highway} connects the internal memory cells of the two neighboring layers, where the only change to the conventional LSTM is to add a highway connection of the last-layer cells to the $c_t$ in Eq. \ref{ct}.
Similar to \cite{zhang2016highway}, \cite{lei2018simple} introduced a simple recurrent unit with highway connections of input $x_t$ on both the memory cells $c_t$ and the output $h_t$ to enhance parallelizable recurrence.
Instead of the shortcut path on an internal memory cell $c_t$, residual LSTM \cite{kim2017residual} was proposed to add a path to the LSTM output $h_t$.
All the above works were motivated by the shortcut and aimed to combine the structure into a specific RNN, but
they focused on building deeper or longer RNNs or enhancing the parallelizability. 
Conformably, limited by the bound $[0,1]$ of the traditional gates, these methods bypassed the essential parts in RNNs: the gate activation, which still seriously suffer from the undertraining problem.
Different from existing explorations, our concentration is the gates.

\section{Methodology}
\subsection{Refined Gate}
The existing \emph{vanilla} saturating gates limit the model learning and mainly result in the \emph{gate undertraining} problem. It further brings the failures of controlling the information flows between hidden states as well as the final weak performance.
Considering the limitations, we propose a new gating mechanism to exploit the power of GRNNs.
Specifically, the refined gate is implemented as a simple shortcut structure by directly short connecting $x_t$ to the outputs of activation function, as shown in Figure \ref{fig:gate} (a).
The refined gate is described as:
\begin{equation} \label{Gate}
\begin{split}%
g_t &= \sigma(\hat{g}_t) \diamond x_t \\
&= \sigma(W_gx_t+U_gh_{t-1}+b_g) \diamond x_t,
\end{split}
\end{equation}
in which $\sigma(\hat{g}_t)$ is the traditional gate function, and $[W_g, U_g, b_g]$ are learnable parameters to control information flows among hidden states.
$\diamond$ denotes the refining operation which can be formulated in the element-wise addition operation $+$ or the element-wise multiplication operation $\times$.
Here, $+$ and $\times$ correspond to two kinds of different refinement operations.
That is, both $x$ and $\sigma$ will obtain the same gradient in $+$ case (as done in residual network). While in the $\times$ case, $x$ and $\sigma$ will separately obtain  gradients with the scales of $\sigma$ and $x$ reversely, treated as the mutual refinement.

{{
Notice, instead of previous gating concept that gating value should be limited in $[0, 1]$, the refined gates are \textbf{boundless}.
It explicitly embraces the decoupled current input features, and differential mapping between the memory state and the input. 
Therefore, the gates can better depict the switch role, which means a more clear ON/OFF controlling on flows, not ambiguous about current input features.
Then the refining mechanism can be directly equipped in GRNNs well as long as it isn't directly applied in hidden state updating such as the Eq. \ref{1e}, \ref{2e} and \ref{3d} due to the gradient explosion \cite{pascanu2013difficulty}, which has been demonstrated in Section 3.2.}

\subsection{GRNNs with Refinement}
\textbf{Refined LSTM }
It is safe to refine the \emph{input} and \emph{output} gates of LSTM, because only learnable parameters themselves (See Eq. \ref{1b} \ref{1c}) are directly affected by the refinement operation in back-propagation.
Therefore, we here modify the input and output gates of traditional LSTM as follows
\begin{subequations} \label{RefLSTM}
\begin{align}
{i}_{t}^\prime &= \sigma(W_ix_{t}+U_ih_{t-1}+b_i)\diamond x_{t}, \label{RefLSTM-i} \\
{o}_{t}^\prime &= \sigma(W_ox_{t}+U_oh_{t-1}+b_o)\diamond x_{t}, \label{RefLSTM-o}
\end{align}
\end{subequations}
where $\diamond$ denotes $+$ or $\times$ operations.
The refinement of gates in Eq. \ref{RefLSTM-i} and Eq. \ref{RefLSTM-o} can be applied independently or jointly.

However, the refinement operation is not suitable for the \emph{forget} gate due to the gradient explosion in memory state learning.
That is, if the \emph{forget} gate is refined, $\delta c_t \rightarrow +\infty$ (shown in Eq. \ref{lstm-forget}).
Then ${f}_t$ will be gradient exploding by referring to $\delta{f}_t = \delta c_t\odot c_{t-1}$, which results in the learning failure.
\begin{equation} \label{lstm-forget}
\delta c_t = \frac{\partial c_T}{\partial c_t} = \prod_{T\ge k \ge t}diag(f_k\diamond x_k).
\end{equation}
Here, $T$ is the index of the last hidden state.

\textbf{Refined GRU }
The refinement on reset gate $r_{t}$ in GRU (defined in Eq. \ref{2b}) is also safe, that is, 
\begin{equation} \label{RefGRU}
{r}_{t}^\prime = \sigma(W_rx_{t}+U_rh_{t-1}+b_r)\diamond x_{t}.
\end{equation}
Similarly, the refinement operation is also not suitable  for the \emph{update} gate due to the gradient explosion between hidden state learning, i.e.,  $\delta h_t \rightarrow +\infty$ because of $\delta h_t = \frac{\partial h_T}{\partial h_t} = \prod_{T\ge k \ge t}diag({z}_k \diamond x_k)$.
Then gradient explosion will directly occur in ${z}_t$ by referring to $\delta{z}_t = \delta h_t\odot (\tilde{h}_t-h_{t-1})$.

\textbf{Refined MGU}
As a variant of GRU, the refinement in MGU (defined in Eq. \ref{3b}) can be also revised as
\begin{equation} \label{RefMGU}
{f}_{t}^\prime = \sigma(W_fx_{t}+U_fh_{t-1}+b_f)\diamond x_{t}.
\end{equation}
Note that, the refinement operation can be just applied into Eq. \ref{3b} while not suitable for Eq. \ref{3d} due to the same gradient explosion problem.



\subsection{Back-propogation of Refining Mechanism}
In order to explore the working mechanism of refined gates, we derive the gradients propagation process in gates to analyze the effectiveness of refined gates in both modes.

Referring to the forward process in Eq. \ref{Gate}, given calculated difference $\delta g_t$ that back-propagates from the following feedforward neural network, we have $\delta \hat{g}_t$, $\delta x_t$, $\delta h_{t-1}$ and $\delta [W_g, U_g, b]$ based on the chain rules, \emph{i.e.},
\begin{subequations} \label{delta-refine}
\begin{align}
&\delta \hat{g}_t = \delta g_t \odot \frac{\partial (\sigma(\hat{g}_t)\diamond x_t)}{\partial \hat{g}_t}  \\
&\delta x_t = \delta g_t \odot \frac{\partial (\sigma(\hat{g}_t)\diamond x_t)}{\partial x_t} \\
&\delta h_{t-1} = \delta \hat{g}_t U , \label{8c}\\
&\delta [W, U, b] = \delta \hat{g}_t [x_t, h_{h-1}, 1]. \label{8d}
\end{align}
\end{subequations}
Since the refining operation $\diamond$ can be represented two forms $+$ and $\times$, each of them generally plays individual roles in gate refining process.

Concretely, for operation $+$, we can reformulate Eq. \ref{delta-refine} as
\begin{subequations} \label{delta-refine+}
\begin{align}
&\delta \hat{g}_t = \delta g_t \odot \hat{g}_t \odot (1-\hat{g}_t), \label{9a}\\
&\delta x_t = \delta \hat{g}_t W + \delta {g_t}, \label{9b}\\
&\delta h_{t-1} = \delta \hat{g}_t U , \label{9c}\\
&\delta [W, U, b] = \delta \hat{g}_t [x_t, h_{h-1}, 1]. \label{9d}
\end{align}
\end{subequations}
We find that GRNN can directly back-propagate $\delta g_t$ to $x_t$, which plays a role like identity mapping \cite{he2016deep} for $x_t$, as shown in Eq. \ref{9b}.
Then the refined gate will relieve the burden on learning $x_t$ and focus on the context relation between hidden states $h_t$.
It is noting that the $+$ refining operation only affects the training of $x_t$, and makes no difference to the back-propagation towards other parts within a recurrent unit such as Eq. \ref{9a}, \ref{9c} and \ref{9d}.

For operation $\times$, we reformulate Eq. \ref{delta-refine} as
\begin{subequations} \label{delta-refinex}
\begin{align}
&\delta \hat{g}_t = \delta g_t \odot \hat{g}_t \odot (1-\hat{g}_t) \times {x_t},  \label{10a}\\
&\delta x_t = \delta \hat{g}_t W + \delta g_t \odot \sigma(\hat{g}_t) , \label{10b}\\
&\delta h_{t-1} = \delta \hat{g}_t U , \label{10c}\\
&\delta [W, U, b] = \delta \hat{g}_t [x_t, h_{h-1}, 1]. \label{10d}
\end{align}
\end{subequations}
Here $x_t$ can be treated as a scaling factor to dynamically adjust amplitudes of activation outputs, as shown in Eq. \ref{10a}.
In fact, the tunable amplitude plays important roles to eliminate the saturation problem of activation functions, as demonstrated in \cite{trentin2001networks}.
Differently, we just use the learnt $x_t$ as the amplitude factor for each neural unit without any additional parameters. 
Apart from the amplitude gains on $x_t$, the residual item $\delta g_t \odot \sigma(\hat{g}_t)$ in Eq. \ref{10b} also help  gates learn their core contents, similar to the + mode.
Besides, it is a slightly larger change than the $+$ refining mode, where the $\times$ operation will affect the back-propagation of the whole recurrent model such as Eq. \ref{10b}, \ref{10c} and \ref{10d}.

\section{Experiments}
In the section, we first explore the proposed architecture in three variants of GRNNs on sequential MNIST \cite{lecun1998gradient} task.
For further illustrating gating functions, we design two independent numerical tasks with qualitative and quantitative analysis on gate controlling ability.
Then we evaluate our refined gate on two general tasks: Language Modeling (LM) and the Scene Text Recognition (STR).
\subsection{Ablation on Sequential Digits} \label{ablation}
\begin{figure}[h]
	\begin{center}
		\includegraphics[width=0.95\linewidth]{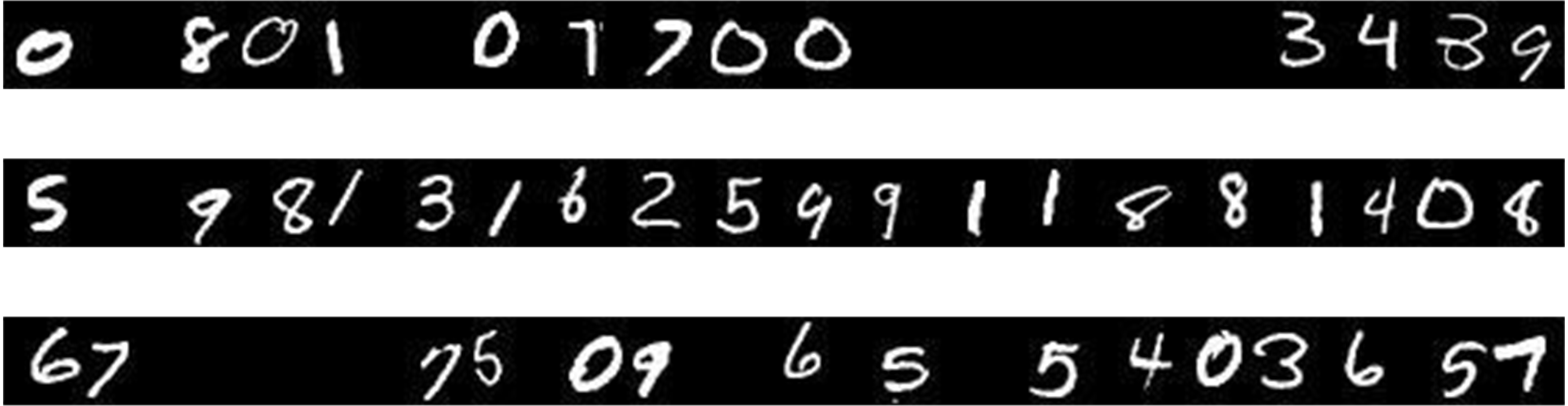}
	\end{center}
	\caption{
		Samples of the sequence recognition task using the MNIST-based dataset.
	}
	\label{fig:ablation}
\end{figure}

\begin{table}[h]
	\caption{Ablation results of refined gates on multiple GRNN architectures. We report the average accuracy and standard deriation of 20 repetitive trails. 
}
	\label{mnist-ablation}
	\centering
	\begin{tabular}{c|c|cc}
		\hline
		Model  & Refined Gate   & Accuracy (+) & Accuracy ($\times$) \\
		\hline
		\multirow{4}{*}{LSTM}
		& --       & 86.17 $\pm$ 0.05 & 86.17 $\pm$ 0.05\\
		& Input    & 86.73 $\pm$ 0.21  & 87.57 $\pm$ 0.20 \\
		& Output     & 87.75 $\pm$ 0.18 & 88.42 $\pm$ 0.20\\
		& Input \& Output     & 87.85 $\pm$ 0.21 & \textbf{88.62 $\pm$ 0.19}\\
		\hline
		\multirow{2}{*}{GRU}
		& --        & 85.22 $\pm$ 0.04  & 85.22 $\pm$ 0.04\\
		& Reset   & 87.70 $\pm$ 0.10   & \textbf{88.09 $\pm$ 0.28}\\
		\hline
		\multirow{2}{*}{MGU}
		& --      & 85.52 $\pm$ 0.10 & 85.52 $\pm$ 0.10 \\
		& Forget   & 88.34 $\pm$ 0.19  & \textbf{88.67 $\pm$ 0.15}\\
		\hline
	\end{tabular}
\end{table}
We construct a more chanllenging dataset from MNIST for evaluation. The sequence recognition images with resolution of 560 x 32 pixels are composed of 12 to 20 digitals randomly selected from digits of original 28 x 28 pixels \cite{lecun1998gradient} with jittering and non-overlapping over all-zero background. 
The training set contains 50,000 images and the test set has 10,000 images.
Figure \ref{fig:ablation} shows three sequential samples.

We train the networks composed of a CNN encoding part with 7-layer convolution layers \cite{shi2016robust}, a single GRNN layer and a connectionist temporal classification \cite{graves2006connectionist} as decoder. Our optimization is using AdaDelta with learning rate 1 for straightforward and convenient evaluation.

Table \ref{mnist-ablation} shows the test results of our refined variants on LSTM, GRU and MGU structures. 
It can be seen that in all three GRNN models, both + and $\times$ refining operations significantly improve the recognition performance, and the $\times$ operation achieves slightly better results than + which might be attributed to the broader activation space of $\times$ as shown in Figure \ref{fig:activation}. 
More details in LSTM, the refined \emph{output} gate (short in RO) outperforms the refined \emph{input} gate (short in RI), because the RO affects input gates in the chain of back-propagation, while the RI can't affect other gates.
Besides, the joint training of RO and RI (short in RIO) can achieve similar results with the RO, or may be slightly improving. 
%
In what follows, we use + as the refining operation by default.
\begin{figure}[h]
	\begin{center}
		\includegraphics[width=1.\linewidth]{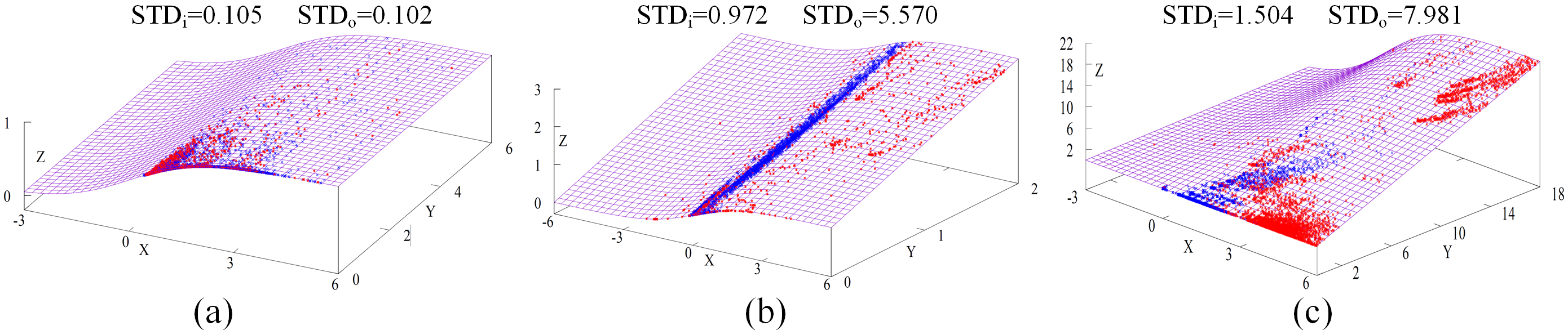}
	\end{center}
	\caption{
		The illustration of activation distribution in gates. (a), (b) and (c) separately denote the distribution of vanilla gates, + and $\times$ refined gates in LSTM. The blue and red points represent the response of \emph{input} and \emph{output} gates. 
	}
	\label{fig:activation}
\end{figure}

\subsection{Ability of Gate Controlling} \label{mini-task}
\subsubsection{ON/OFF Indication on Adding Task}

Adding problems \cite{hochreiter1997long} were once designed for evaluating long-term memorizing ability of RNN variants. While we focus on the gate controlling power, we make a minor revision of the task. 
The inputs have two binary sequences with a pre-set length $L\in\{10, 20, 50, 500, 5000, 50000\}$. For example, given the sample `0110000000' and `0100100000' as two inputs, the corresponding annotation is denoted as `0001100000'.
Note that the addition number are binary sequences in a reverse order. Since in a unidirectional RNN state learning setting, the current output depends on the information of the last state (i.e., the carry of each single bit addition is added to the next bit).

For each input length, we randomly generate a training set of 10,000 binary sequences and a test set of 5,000 samples, respectively. Thus, \textit{we expect the gates to turn ON/OFF for indicating the addition information, \textit{i.e.}, 0/1 carry bit. }
We evaluate effects of gates by feeding feature sequence into a single-direction RNN layer with 4 hidden states, after that following a fully connected layer to transform the hidden state vector as a sequential 2-class predictor.

As the adding problem is the simple addition operation, once the model converged, the predict accuracy tends to be 100\%. 
Therefore, we compare the number of the completely converge epoches instead of the prediction accuracy to evaluate GRNNs as shown in Table \ref{mini-add}.
 \begin{table}[h]
	\caption{The epoches of complete converge according to $L$. $\infty$ means non-convergence. `K' means thousand. Lower values represents faster and better training. }
	\label{mini-add}
	\centering
	\scalebox{0.84}{
		\begin{tabular}{ccccccc}
			\hline
			Architecture & Mode & L=10 & L=20 & L=50 & L=100 & L=500 \\
			\hline
			LSTM  & None & 15 &  32  & 92    & $\infty$    & $\infty$\\		
			LSTM-RI & + & 6  & 6 & 6  & 12    & $\infty$ \\
			LSTM-RI & $\times$  & 13  & 18 & 26  & 57    & $\infty$ \\
			LSTM-RO &  + & 6 & 6 & 6   & 12 & 12\\
			LSTM-RO &  $\times$ & 16 & 16 & 16   & 16 & 25\\		
			\hline
			GRU & None & 23  & 60  & $\infty$    & $\infty$ &  $\infty$ \\
			GRU-RR  &  + & 22    & 22 & 68    & $\infty$ &  $\infty$\\
			GRU-RR  & $\times$ & 17    & 29 & 140    & $\infty$ &  $\infty$ \\
			\hline
			MGU & None & 23  & 92 & $\infty$    & $\infty$ &  $\infty$ \\
			MGU-RF  &  + & 21  & 23 & 66   & $\infty$ &  $\infty$ \\
			MGU-RF  &  $\times$ & 21  & 29 & 91  & $\infty$ &  $\infty$\\
			\hline
		\end{tabular}
	}
\end{table}
\begin{figure}[h]
	\begin{center}
		\includegraphics[width=1.\linewidth]{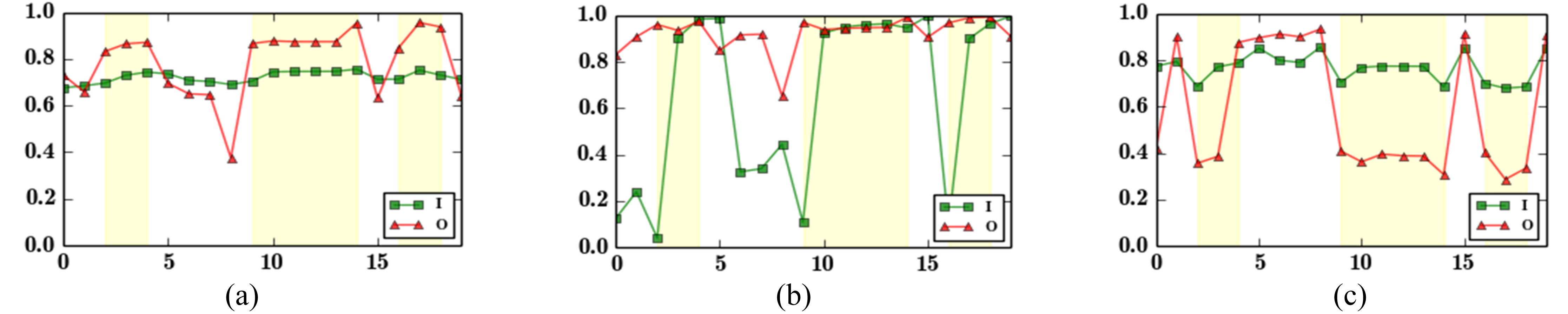}
	\end{center}
	\caption{
		The adding process in LSTM. (a), (b) and (c) separately denote the activation values of $\sigma$ in LSTM, LSTM-RI and LSTM-RO in + refining mode along the binary sequence. The green and red points represent the response of \emph{input} and \emph{output} gating functions. The yellow regions mean carry bit regions.
	}
	\label{fig:add}
\end{figure}
\begin{figure}[h]
	\begin{center}
		\includegraphics[width=1.\linewidth]{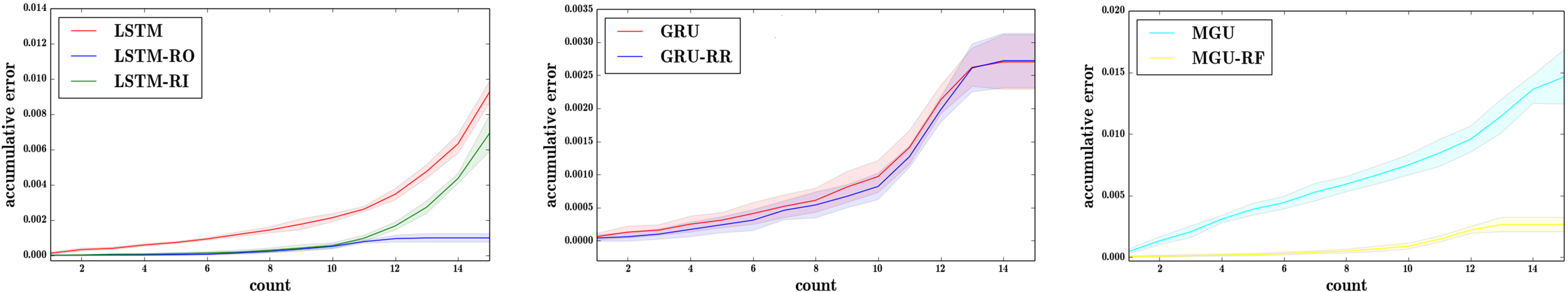}
	\end{center}
	\caption{Distribution of accumulative error rates on different GRNNs with + refined gates. The horizontal and vertical axis represent the count value and the accumulative error rate, respectively.
	}
	\label{fig:count}
\end{figure}
\begin{figure}[h]
	\begin{center}
		\includegraphics[width=1.\linewidth]{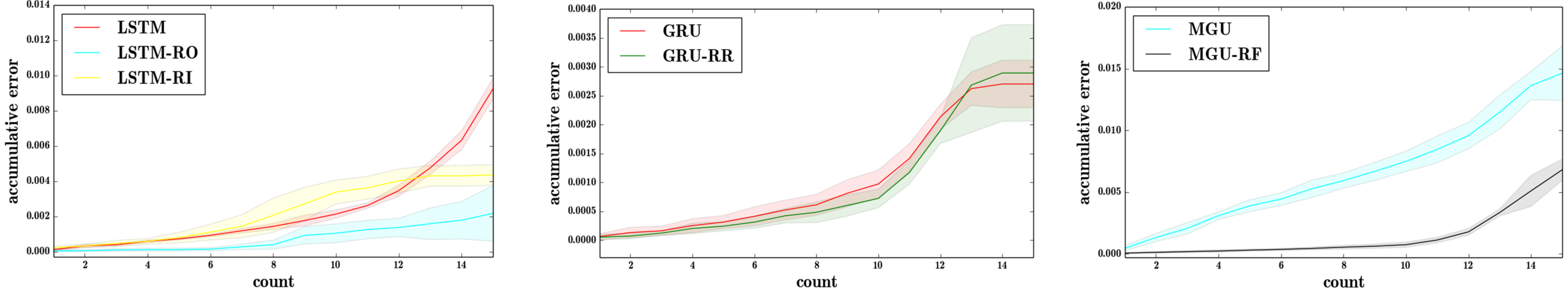}
	\end{center}
	\caption{
		Distribution of accumulative error rate on different GRNNs with $\times$ refined gates. The horizontal and vertical axis represent the count value and the accumulative error rate, respectively.
	}
	\label{fig:count-error}
\end{figure}
Table \ref{mini-add} shows all listed GRNN models can converge when $L$ equals to 10 or 20, and LSTM can hold longer.
Results can be quantatively evaluated in two aspects: \textbf{converging speed} and \textbf{input length}.
It costs more epoches for converging with the input length increasing, and  the refined RNN models obviously obtain better performance than their vanilla structures.
It worth noting that the LSTM-RO remarkably converges rapidly even when L=50,000 (not listed in the table) while vanilla models can only hold less than L=50, which illustrates the prominent superior performance of LSTM-RO comparing to the others (holding \textit{1000 times longer} input length with less training epoches). {We also speculate that the faster converge of the + than $\times$ mode is due to the intrinsic context in this task, where the mathematical carry bit can be more easily obtained through residual operations.} 
Note that we mention input length here only for evaluating the ON/OFF gate roles in different cases.
{
In other words, even with much longer sequences, the goal of the networks is always the same, to learn the \textit{carry bit}, which requires only the previous one memory state and the current input addition numbers. 
As for the controlling of long-dependency flows, we would go further to discuss in next section.}

From the qualitative perspective of the \textbf{gate controlling ability}, we attribute the essence of the sequential learning in the adding task to indicating the carry bit.
Conformably, we find that the refining operation makes gates more sensitive to the carry bit, when noticing the different amplitude of activation values in the gates, as shown in Figure \ref{fig:add}.
For detail, the \emph{input} gate curve in LSTM is quite steady as a saboteur (Figure \ref{fig:add} (a)), while the input curve of LSTM-RI has drastic change clearly indicating each carry bit (Figure \ref{fig:add} (b)). Similarly, LSTM-RO has a very sensitive activation in its output gate (Figure \ref{fig:add} (c)). The adding task verifies the refined gates are more likely to perceive the gating roles of the adding problem, which shows that the refining operation can help eliminate the \emph{gate} \emph{undertraining}.

\subsubsection{Dependency Controlling in Counting Task}
We further evaluate the dependency controlling of gate in the counting task. The input is a binary sequence with length of 20. Each binary digit (0 or 1) in input sequences is encoded into a 2-dimensional one-hot encoding.
We evaluate effects of gates by feeding feature sequences into an LSTM with 2 hidden states, after that following a fully connected layer to predict the repetition number of the last appearing digit.
For instance, given the input sequence ended with `11110010110100100000', the predict value should be 5.
\begin{table}[h]
	\caption{Perplexity results of different recurrent models on the word-level PTB task. Lower is better.}
	\label{ptb}
	\centering
	\begin{tabular}{l|cc}
		\hline
		Model  		& Validation & Test \\
		\hline
		\multicolumn{3}{l}{\textbf{SoA Reported:}}	 \\
		\hline
		Unregularized LSTM \cite{zaremba2014recurrent} & 119.4  	 &  115.6 \\
		Regularized LSTM \cite{zaremba2014recurrent}	 	& 86.2  	 &  82.7 \\
		Noisy LSTM+ NAH	 \cite{gulcehre2016noisy}		& 111.7 	 &  108.0 \\
		Variational LSTM \cite{gal2015a}		& 81.9  	 &  79.7 \\
		Zoneout    \cite{krueger2016zoneout:}	  	 		& 84.4 	 	 &  80.6 \\
		QRNN \cite{bradbury2016quasirecurrent}				& 82.9 	 	 &  79.9 \\
		\hline
		\hline
		\multicolumn{3}{l}{\textbf{Our Setup:}}	 \\
		\hline	
		LSTM \cite{zaremba2014recurrent} & 85.9 	     &  82.1 \\
		LSTM-RI (Ours)  	& 82.6 	     &  80.1 \\
		LSTM-RO (Ours)		& {82.3} 	&  {79.6} \\
		LSTM-RIO (Ours)  	& \textbf{82.1} &  \textbf{79.5} \\
		\hline		
	\end{tabular}
\end{table}
In this setting, gates are responsible for controlling information flow related to the repetition number of the last appearing digit. 
Figure \ref{fig:count} and \ref{fig:count-error} shows the accumulative error rates along different count ranges, in which the refined gates effectively decrease the error rates, especially in predicting larger counting number.
We attribute it to improvement on the gating power of controlling the long-dependency flows.
Similarly, as mentioned in adding task above (See Table \ref{mini-add}), the refined RNN shows obvious improvement on the long-dependence sequential flows (holding over 1000 times longer input length with LSTM-RO compared with vanilla LSTM structure).
\begin{table*}[thb]
	\centering
	\caption{Perplexity statistics of three GRNNs (\emph{i.e.}, LSTM, GRU and MGU) and their {refined} variants on the PTB word-level prediction. Lower is better.}
	\label{ptb-2}
	\begin{tabular}{c|cccc|cc|cc}
		\hline
		Param  &	\multicolumn{4}{c|}{LSTM} & \multicolumn{2}{c|}{GRU} & \multicolumn{2}{c}{MGU} \\
		$\#$Hidden & LSTM & LSTM-RI & LSTM-RO & LSTM-RIO & GRU & GRU-RR & MGU & MGU-RF \\ \hline
		200 & 106.2 & 104.5 & 103.2 & \textbf{103.1} & 108.2 & \textbf{107.8} & 113.1 & \textbf{111.4} \\
		400 & 89.3 & 86.7 & \textbf{86.4} & \textbf{86.4} & 93.1 & \textbf{92.2} & 96.0 & \textbf{94.1} \\
		600 & 86.2 & 83.8 & 84.2 & \textbf{83.7} & 92.6 & \textbf{90.8} & 95.2 & \textbf{92.2} \\
		800 & 86.7 & \textbf{83.6} & 84.0 & \textbf{83.6} & 93.3 & \textbf{92.0} & 94.8 & \textbf{91.5} \\
		\hline
	\end{tabular}	
\end{table*}
\begin{table}[h]
	\caption{Test BPCs of models on the char-level PTB task. Lower is better.}
	\label{cptb}
	\centering
	\begin{tabular}{l|cc|c}
		\hline
		Model  		& $\#$Layer & $\#$Step & Test BPC \\
		\hline
		\multicolumn{4}{l}{\textbf{SoA Reported:}}	 \\
		\hline
		RNN \cite{neyshabur2016path-normalized}  				 & 1   &  150 & 2.89\\
		LSTM \cite{zaremba2014recurrent} & 3  	 &  150 & 1.48\\
		EURNN \cite{henaff2016recurrent}		 	 & 1  	 &  150 & 1.69\\
		HM-LSTM \cite{Ha2016HyperNetworks}			 & 3  	 &  100 & 1.30 \\
		IndRNN \cite{li2018independently} 	 	     & 6  	 &  50  & 1.26\\
		IndRNN \cite{li2018independently}	  	 	 & 21 	 &  50  & 1.21\\
		NRU    \cite{chandar2019towards}	  	 	 & 1     &  150 & 1.47\\
		nnRNN  \cite{kerg2019non-normal}		 & 3 	 &  150 & 1.47\\
		\hline
		\hline
		\multicolumn{4}{l}{\textbf{Our Setup:}}	 \\
		\hline
		RNN	 				& 1 	     &  50  & 1.49\\
        Residual LSTM \cite{kim2017residual} &3  & 50    &1.37\\
		IndRNN\footnotemark[1] \cite{li2018independently}  	& 3 	     &  50  & 1.36\\
		NRU\footnotemark[2] \cite{chandar2019towards}  		& 3 	     &  50  & 1.37\\	
		\hline
		LSTM     			& 1 	     &  50 & 1.46\\	
		LSTM-RI (Ours)     	& 1 	     &  50 & 1.41\\
		LSTM-RO (Ours)     	& 1 	     &  50 & \textbf{1.39}\\
		LSTM-RIO (Ours)		& 1			 &  50 & \textbf{1.39}\\
		\hline
		LSTM     			& 3 	     &  50 & 1.38\\
		LSTM-RI (Ours)     	& 3 	     &  50 & 1.32\\
		LSTM-RO (Ours)     	& 3 	     &  50 & 1.31\\
		LSTM-RIO (Ours)		& 3			 &  50 & \textbf{1.29}\\
		\hline		
	\end{tabular}
\end{table}
\begin{table}[h!]
	\caption{Validation and test BPCs of diffenrent recurrent models on EnWik8 dataset. Lower is better.}
	\label{enwik}
	\centering
	\begin{tabular}{l|cc|cc}
		\hline
		Model  		& $\#$Layer & $\#$Step & Valid & Test \\
		\hline
		\multicolumn{4}{l}{\textbf{SoA Reported:}}	 \\
		\hline
		LSTM \cite{zaremba2014recurrent}  	 	& 3  & 100 & -- & 1.67\\
		Grid-LSTM \cite{zaremba2014recurrent} 	& 3  & 100 & -- & 1.58\\
		MI-LSTM \cite{wu2016on}	   			& 3  & 100 & -- & 1.44\\
		LN LSTM \cite{Ba2016Layer} 			& 3  & 100 & -- & 1.46 \\
		HM-LSTM \cite{Ha2016HyperNetworks}                  		& 3  & 100 & -- & 1.40 \\
		HyperLSTM \cite{Ha2016HyperNetworks} 			 	  		& 3  & 100 & -- & 1.39 \\
		QRNN \cite{bradbury2016quasirecurrent} 		 & 4  	 &  200 & -- & 1.33\\
		SRU \cite{lei2018simple}	  	 	 & 6 	 &  100 &1.29& 1.30\\
		\hline
		\hline
		\multicolumn{4}{l}{\textbf{Our Setup:}}	 \\
		\hline
        Residual LSTM \cite{kim2017residual} &3          &  100 & 1.39& 1.40\\	
		QRNN(k=1)\footnotemark[3] \cite{bradbury2016quasirecurrent}  		& 3 	     &  100 & 1.39& 1.40\\	
		SRU\footnotemark[4] \cite{lei2018simple}      	& 3 	     &  100 & 1.36& 1.38\\
		\hline
		LSTM   	& 3 	     &  100 & 1.38 & 1.40  \\
		LSTM-RI(Ours)     	& 3 	     &  100 & 1.37& 1.39\\
		LSTM-RO(Ours)     	& 3 	     &  100 & \textbf{1.36}& 1.38\\
		LSTM-RIO(Ours)     	& 3 	     &  100 & \textbf{1.36}& \textbf{1.37}\\
		\hline		
	\end{tabular}
\end{table}
\subsection{Language Modeling}
\subsubsection{Word-level PennTree Bank}
The Penn TreeBank (PTB) dataset \cite{marcus1993building} provides data for language modeling. There are 10,000 words in the vocabulary, including 929K words in the training set, 73K in the validation set, and 82K in the test set.
\begin{table*}[thb]
	\caption{Recognition accuracies on the 5 general recognition datasets in a lexicon-free setting. For more straightforward evalutaion, we compute the average (AVG) accuracies of the 5 benchmark results.}
	\label{textRecog}
	\centering
	\begin{tabular}{l|ccc|ccccc|c}
		\hline
		Method          	& $\#$Layer&$\#$Step &Type & III5K & SVT  & IC03  & IC13 & IC15 & AVG\\
		\hline
		\multicolumn{7}{l}{\textbf{SoA Reported:}}	 \\
		\hline
		FAN \cite{cheng2017focus} & 1& 65 &-- & 83.7 & 82.2 & 91.5 & 89.4     &  63.3 & 82.0\\
		ASTER \cite{shi2018aster} & 2& 26 & -- & {91.9} & {88.8} & {93.5} & 89.8     &  -- & --\\
		MORAN \cite{luo2019moran} & 1& 26 & -- & 84.2 & 82.2 & 91.0 & 90.1     &  {65.6} & 82.6\\
		\hline
		\hline
		\multicolumn{4}{l}{\textbf{Our Setup:}}	 \\
		\hline
		Baseline  &0	  & 65 & --   & 86.1 & 84.7 & 92.6 & 90.3 & 67.4 & 84.2  \\
		\hline
		LSTM  &1	  & 65 & --   & 86.2 & {85.5} & 93.8 & 90.8 & 67.4 & 84.7(+0.5)  \\
		LSTM-RO (Ours) &1 & 65 & Only Encoder & 86.2 & 84.9 & 93.9 & 91.5 & 68.9 & 85.1(+0.9)    \\
		LSTM-RO (Ours)   &1 & 65 & Both 		   & {86.6} & {85.5}& {94.3} & {91.9} & {69.0} & {85.5(+1.3)}\\
		\hline
		LSTM   &2 & 65 & --   & 86.8 & 84.1 & 93.1     & 91.5 &  67.8 &  84.6(+0.4)  \\
		LSTM-RO (Ours)  &2 & 65 & Only Encoder & 87.2 & 85.3 & 93.5  & 92.0 &  69.2 &  85.4(+1.2)   \\
		LSTM-RO (Ours)  &2 & 65 & Both & \textbf{87.7} & \textbf{86.1} & \textbf{93.8} & \textbf{92.1} &  \textbf{69.7} & \textbf{85.9(+1.7)}\\
		\hline
	\end{tabular}
\end{table*}
Our experimental settings follow the standard setup \cite{zaremba2014recurrent}: 2 layers of 650 units in each layer with the sequence length of 35 as the encoder. Then a fully connected layer predicts one of the 10,000 words. The minibatch size is 20. We apply 50$\%$ dropout on the non-recurrent connections. We train for 39 epoches with a learning rate of 1 and then reduce it by a factor of 1.2 per epoch.
{
Note that state-of-the-art (SoA) methods on the task explore the network design or the training strategies, while we focus on the innermost gating mechanism. As fancy techniques can bring improvements on overall performance but do no help to our fair comparison on the gates themselves,
we compare our proposed model in the consistent setting (\emph{i.e.}, 2 layers of 650 or 640 recurrent units), as shown in Table \ref{ptb}.}
For fair comparison, we implement the Regularized LSTM \cite{zaremba2014recurrent} as baseline in our setups. We also evaluates in Table \ref{ptb-2} on the refined model with LSTM, GRU and MGU over various hidden units for comprehensive verification.
It clearly shows that all the refined gates achieve improvements.

\subsubsection{Char-level PennTree Bank}
The character-level PennTree Bank (cPTB) task is to predict the next character in a text corpous at every character position, given all previous text. Our experiments comply with the setting of early work \cite{chung2016hierarchical}. The model consists of an input embedding layer, an RNN module and an output layer. The RNN module has three layers, and the whole models are trained sequentially on the entire corpus, splitting it into sequences of length 50 for truncated back-propagation through time. The minibatch size is 128. We train models using Adam with an initial learning rate of 0.0002 and drop by a factor of 5 with patience 20.

We compare the performance of different models on the task in Table \ref{cptb} in terms of test mean bits per character (BPC), where lower value indicates better results. Since reported SoAs are trained under different settings, \emph{e.g.}, recurrent layers can be 1 \cite{kerg2019non-normal} or 21 \cite{li2018independently},
{
we implement the models in the same setup for direct evaluation and focus on the improvement on the same module setting with 1 and 3 layers.}
{It is noting that existing variants \cite{li2018independently} motivated to improve gradient propagation among RNN layers, while our method is focused on the roles of gating mechanism, more similar with the work \cite{chandar2019towards} which largely falls behind our method.
It shows the improvement of our refined LSTM can be further enlarged in a 3-layer settting.}
\subsubsection{Wikipedia}
We use EnWik8\footnote[1]{The human knowledge compression contest 2012. http://prize.hutter1.net/.}, a large public dataset for character-level language modeling. Following standard practice, we use the first 90M characters for training and the remaining 10M split evenly for validation and test. Similar to former work \cite{zaremba2014recurrent}, we use a batch size of 128 with the unroll size of 100 for truncated backpropagation during training. We use the Adam optimizer and the same learning rate scheduling following \cite{lei2018simple} with a maximum of 100 epoches (about 700,000 iterations).
The existing models shows close BPCs on tasks even in large variants of network settings (\emph{e.g.}, the number of the recurrent layer, the hidden units in a wide range). Table \ref{enwik} shows the improvement on the validation and test evaluation under the same settings with existing methods.

\subsection{Scene Text Recognition} \label{text-exp}
In addition to language modeling, we evalutate on visual data in text recognition tasks. Following the common settings, we use 2 synthetic datasets MJSynth \cite{jaderberg2014synthetic} and SynText \cite{Gupta16} as our training set, and 5 general text benchmarks (including \emph{III5K}, \emph{SVT}, \emph{IC03}, \emph{IC13} and \emph{IC15} \cite{baek2019STRcomparisons}) as evaluation sets without any extra finetuning.
We use the ResNet-32 \cite{cheng2017focus} as backbone, sequence modeling (Bidirectional LSTM) for a contextural feature, and an attention-based decoder (in which one LSTM layer is applied for sequential decoding).
Here in the sequence modeling, we choose the LSTM with refined output gate for evaluation.
We train the model as: (1) backbone without sequence modeling as baseline, (2) traditional LSTM without refined gates, (3)  LSTM-RO in its encoder and (4) LSTM-RO in both encoder and decoder.

Table \ref{textRecog} shows the recognition accuracies on the 5 benchmarks.
For more convincing evaluation, we show the results of 1 layer and 2 layers of BiLSTMs in the top of its encoder.
Results show LSTM-RO in encoder can outperform the baseline, and LSTM-RO in both the encoder and decoder get the best performance.
We also compare our results with three strong baseline\cite{cheng2017focus,luo2019moran,shi2018aster} and achieve competitive results, but only falls behind \cite{shi2018aster} on III5K and SVT because \cite{shi2018aster} used the deeper 45-layers ResNet as feature encoder.
Intrinsically, the 2-layer LSTM-RO further enlarges the improvement above the baseline results from 1.3\% to 1.7\%, largely surpassing the traditional LSTM gain of 0.4\%.

{Note that the comprehensive analysis of modules in recent work \cite{baek2019STRcomparisons} shows the 2-layer BiLSTM improve the baseline model (without the whole sequence modeling modules) by only slight gain in average (only 0.5\% improvement in the model with vanilla LSTM layers in our setup), the \textbf{1.7\% improvement with only minor revision} above the dominated sequence modeling method (LSTM) without extra cost is considerably significant in the STR tasks.}
$\footnotetext[1]{The SoA models are trained in different settings, and can not be directly comparable in the unified setups. We incorporated the IndRNN module with its released code at
	https://github.com/Sunnydreamrain/IndRNN$\_$Theano$\_$Lasagne for evaluation in the same setting. And NRU, QRNN and SRU are similar.}$
$\footnotetext[2]{We use the code at https://github.com/apsarath/NRU.}$
$\footnotetext[3]{We use the code at https://github.com/salesforce/pytorch-qrnn.}$
$\footnotetext[4]{We use the code at https://github.com/taolei87/sru.}$

\section{Conclusion}
In this paper, we propose the a new gating mechanism for handling the \emph{gate} \emph{undertraining} problem.
The refining operation is implemented by directly short connecting the inputs to the outputs of activating function, which can effectively enhance the controlling ability of gates.
We illustrate the refined gates, offer a deeper understanding of the gating mechanism, and then evaluate their performance on several popular GRNNs.
Extensive experiments on various sequence data including the scene text recognition, language modeling and arithmetic tasks,
demonstrate the effectiveness of the refining mechanism.
In future, we will further explore the gating mechanism in GRNNs.



\bibliographystyle{ACM-Reference-Format}
\bibliography{egbib}

\end{document}